\documentclass{article}

\usepackage{arxiv}

\usepackage[utf8]{inputenc} 
\usepackage[T1]{fontenc}    
\usepackage{hyperref}       
\usepackage{url}            
\usepackage{booktabs}       
\usepackage{amsfonts}       
\usepackage{nicefrac}       
\usepackage{microtype}      
\usepackage{lipsum}		
\usepackage{graphicx}
\usepackage{natbib}
\usepackage{doi}
\usepackage{amsmath,amssymb,amsfonts}
\usepackage{algorithmic}
\usepackage{graphicx}
\usepackage{textcomp}
\usepackage[ruled]{algorithm2e}
\usepackage{bbold}
\usepackage{graphics}
\usepackage{multirow}
\usepackage{hyperref}

\usepackage{subcaption}
\usepackage{booktabs}
\usepackage{placeins}
\usepackage{cite}
\usepackage{amsmath,amssymb,amsfonts}
\usepackage{algorithmic}
\usepackage{algorithmic}
\usepackage[ruled]{algorithm2e}
\usepackage{graphicx}
\usepackage{textcomp}
\usepackage{float}
\usepackage{multirow}

\title{Antithetic Riemannian Manifold And Quantum-Inspired Hamiltonian Monte Carlo}

\author{ {Wilson Tsakane Mongwe} \\
	School of Electrical Engineering\\ University of Johannesburg\\
	Auckland Park, 2006, South Africa\\
	\texttt{wilsonmongwe@gmail.com} \\
	\And
	
	{Rendani Mbuvha} \\
School of Statistics and Actuarial Science\\ University of Witwatersrand\\
Johannesburg, 2000, South Africa\\
\texttt{rendani.mbuvha@wits.ac.za} \\
	\And
	{Tshilidzi Marwala} \\
	School of Electrical Engineering\\ University of Johannesburg\\
	Auckland Park, 2006, South Africa\\
	\texttt{tmarwala@uj.ac.za} \\
}

\renewcommand{\headeright}{}
\renewcommand{\undertitle}{}
\renewcommand{\shorttitle}{Antithetic Riemannian Manifold And Quantum-Inspired Hamiltonian Monte Carlo}

\hypersetup{
pdftitle={A template for the arxiv style},
pdfsubject={q-bio.NC, q-bio.QM},
pdfauthor={David S.~Hippocampus, Elias D.~Striatum},
pdfkeywords={First keyword, Second keyword, More},
}

\begin{document}
\maketitle

\begin{abstract}
Markov Chain Monte Carlo inference of target posterior distributions in machine learning is predominately conducted via Hamiltonian Monte Carlo and its variants. This is due to Hamiltonian Monte Carlo based samplers ability to suppress random-walk behaviour. As with other Markov Chain Monte Carlo methods, Hamiltonian Monte Carlo produces auto-correlated samples which results in high variance in the estimators, and low effective sample size rates in the generated samples. Adding antithetic sampling to Hamiltonian Monte Carlo has been previously shown to produce higher effective sample rates compared to vanilla Hamiltonian Monte Carlo. In this paper, we present new algorithms which are antithetic versions of Riemannian Manifold Hamiltonian Monte Carlo and Quantum-Inspired Hamiltonian Monte Carlo. The Riemannian Manifold Hamiltonian Monte Carlo algorithm improves on  Hamiltonian Monte Carlo by taking into account the local geometry of the target, which is beneficial for target densities that may exhibit strong correlations in the parameters. Quantum-Inspired Hamiltonian Monte Carlo is based on quantum particles that can have random mass. Quantum-Inspired Hamiltonian Monte Carlo uses a random mass matrix which results in better sampling than Hamiltonian Monte Carlo on spiky and multi-modal distributions such as jump diffusion processes. The analysis is performed on jump diffusion process using real world financial market data,  as well as on real world benchmark classification tasks using Bayesian logistic regression.  The results show that the new antithetic Riemannian Manifold Hamiltonian Monte Carlo and Quantum-Inspired Hamiltonian Monte Carlo algorithms produce higher effective sample size rates than antithetic Hamiltonian Monte Carlo. Furthermore,  antithetic Quantum-Inspired Hamiltonian Monte Carlo produces the highest effective sample sizes on the jump diffusion datasets, while Riemannian Manifold Hamiltonian Monte Carlo produces the highest effective sample sizes on the logistic regression datasets.
\end{abstract}


\section{Introduction}

Coupling of Markov chains has been relied upon to prove the convergence of Markov Chain Monte Carlo algorithms (MCMC), as well as for providing lower bounds for the effective sample sizes generated by MCMC methods \citep{rosenthal1997faithful,johnson1998coupling, johnson1996studying, jacob2017unbiased, bou2020coupling}. Theoretical results in recent times show that two Hamiltonian Monte Carlo chains can be coupled by giving them the same set of random numbers \citep{heng2019unbiased, bou2020coupling,piponi2020hamiltonian}. In other words, even though the chains might be initialised with different states, their dynamics will at some point  become indistinguishable \citep{bou2020coupling,piponi2020hamiltonian}. Markov chain coupling theory has also been used to provide unbiased Hamiltonian Monte Carlo estimators \citep{glynn2014exact,jacob2017unbiased, heng2019unbiased}. An approach of constructing a pair of HMC chains, where the momentum variable is shared,  such that they create unbiased chains is presented by \citet{heng2019unbiased}. On the other hand, \citet{bou2020coupling} propose a new approach called contractive sampling where the momentum variable is not shared between the coupled chains. 

\citet{piponi2020hamiltonian} use approximate coupling theory to construct the antithetic Hamiltonian Monte Carlo algorithm. These anti-correlated chains are created by running the second chain with the momentum auxiliary variable having the opposite sign of the momentum of the first chain \citep{piponi2020hamiltonian, mongwe2020antithetic}. Their results show that adding antithetic sampling to HMC increases the effective sample size rates. \citet{mongwe2020antithetic} present the antithetic versions of Magnetic Hamiltonian Monte Carlo \citep{tripuraneni2017magnetic}, Magnetic Momentum Monte Carlo \citep{brofos2020non} and Separable Shadow Hamiltonian Hybrid Monte Carlo \citep{sweet2009separable}. Their analysis was performed on Bayesian logistic regression and Bayesian neural networks. Their results also support that adding antithetic sampling to the base algorithm increases the effective sample sizes, even after accounting for the execution time. In this work we expand on this by providing new algorithms which are antithetic versions of Riemannian Manifold Hamiltonian Monte Carlo \citep{girolami2011riemann} and Quantum-Inspired Hamiltonian Monte Carlo \citep{liu2019quantum}.  We compare the performance of these new algorithms against the antithetic Hamiltonian Monte Carlo method.

The Riemannian Manifold Hamiltonian Monte Carlo (RMHMC) algorithm improves on vanilla Hamiltonian Monte Carlo by taking into account the local geometry of the target \citep{girolami2011riemann,cobb2019introducing}. This is particularly important for ill conditioned targets \citep{girolami2011riemann,cobb2019introducing, liu2019quantum}. This results in samples with very low auto-correlations and consequently large effective sample sizes. The main disadvantage of RMHMC is that the Hamiltonian is non-separable, thus leading to a computationally expensive implicit integration scheme being required \citep{cobb2019introducing, girolami2011riemann}. We  leverage the benefits of RMHMC over HMC and antithetic sampling to create the antithetic RMHMC sampler that should produce higher effective samples than antithetic HMC.

Quantum-Inspired Hamiltonian Monte Carlo (QIHMC) uses a random mass matrix which results in better sampling than HMC and RMHMC on spiky and multi-modal distributions \citep{liu2019quantum}. QIHMC achieves the outperformance over HMC without any noticeable increase in the execution time. The main drawback of QIHMC is the requirement for the user to specify the distribution of the mass matrix \citep{liu2019quantum}. In this work, we rely on the outperformance of QIHMC over HMC to create antithetic QIHMC which should result in higher effective sample sizes when compared to HMC. 

The analysis is performed using real world financial market data on the Bayesian inference of jump diffusion processes, as well as on real world benchmark classification tasks using Bayesian logistic regression. The results show that the new antithetic Riemannian Manifold Hamiltonian Monte Carlo and Quantum-Inspired Hamiltonian Monte Carlo algorithms produce higher effective sample size rates than antithetic Hamiltonian Monte Carlo. The antithetic Riemannian Manifold Hamiltonian Monte Carlo method produces the largest effective sample sizes on the logistic regression datasets. Furthermore, antithetic Quantum-Inspired Hamiltonian Monte Carlo produces the highest effective sample size rates normalised by execution time of all the methods, with significant outperformance on the jump-diffusion datasets. 

\textbf{Contributions:} The main contributions of this work are as follows:
\begin{itemize}
    \item We present two new algorithms being antithetic Riemannian Manifold (A-RMHMC) and antithetic Quantum-Inspired Hamiltonian Monte Carlo (A-QIHMC) 
    \item We provide the first application of antithetic MCMC algorithms to the Bayesian inference of jump diffusion processes in financial markets.
\end{itemize}

The remainder of this paper is structured as follows: Sections \ref{sec:hmc} to \ref{sec:rmhmc} discuss the HMC, RMHMC and QIHMC algorithms on which the new antithetic algorithms are based, Section \ref{sec:anti} presents the two new antithetic algorithms that we are proposing along with the antithetic HMC method, Section \ref{sec:exp} outlines the experiments conducted, Section \ref{sec:results} presents and discusses the results of the experiments and we provide the conclusion in Section \ref{sec:conclusion}.

\section{Hamiltonian Monte Carlo}
\label{sec:hmc}

Hamiltonian Monte Carlo (HMC) uses first order gradient information of the target posterior to suppress random walk behaviour of the Metropolis-Hastings \citep{duane1987hybrid, neal2011mcmc}. An auxiliary momentum variable $\mathbf{p}$ is added to the parameter space. The  Hamiltonian $ {\rm H} (\mathbf{w},\mathbf{p})$ is written as follows \citep{neal1993bayesian}:
\begin{equation}
\label{eq:ham}
    {\rm H}(\mathbf{w},\mathbf{p})= {\rm U}(\mathbf{w})+ {\rm K} (\mathbf{p})
\end{equation}
where $ {\rm U} (\mathbf{w})$ is the negative log-likelihood of the target posterior distribution and ${\rm K} (\mathbf{p})$ is the kinetic energy defined by the kernel of a Gaussian with a covariance mass matrix $\mathbf{M}$ \citep{neal2012bayesian}:
\begin{equation}
    {\rm K} (\mathbf{p}) =\frac{1}{2}\text{log}\left( (2\pi)^D |\mathbf{M}|\right) + \frac{\mathbf{p}^{\rm T} \mathbf{M}^{-1}\mathbf{p}}{2}.
\end{equation}
The trajectory vector field is defined by considering the parameter space as a physical system that follows Hamiltonian dynamics \citep{neal2011mcmc}. The equations governing the trajectory of the chain are then defined by Hamilton's equations at a fictitious time $t$ as follows \citep{neal1993bayesian}:
\begin{equation}
    \frac{{\rm d} \mathbf{w}}{\partial t} =  \frac{\partial {\rm H}(\mathbf{w},\mathbf{p})}{\partial \mathbf{p}} ;
\quad \frac{{\rm d} \mathbf{p}}{\partial t} =  -\frac{\partial {\rm H}(\mathbf{w},\mathbf{p})}{\partial \mathbf{w}}.
\end{equation}

The Hamiltonian in equation \eqref{eq:ham} is separable, thus the the leapfrog integrator can be used to traverse the parameter space \citep{duane1987hybrid, neal1993bayesian}. The update equations for the leapfrog integration scheme are \citep{duane1987hybrid, neal2012bayesian}:
\begin{equation}
\label{eq:leapfrog}
\begin{split}
    \mathbf{p}_{t+\frac{\epsilon}{2}} &= \mathbf{p}_{t} + \frac{\epsilon}{2}\frac{\partial H \left(\mathbf{w}_{t}, \mathbf{p}_{t}\right)}{\partial \mathbf{w}} 
    \\
   \mathbf{w}_{t+\epsilon} &= \mathbf{w}_{t} + \epsilon\mathbf{M}^{-1}\mathbf{p}_{t+\frac{\epsilon}{2}}
\\
    \mathbf{p}_{t+\epsilon} &= \mathbf{p}_{t+\frac{\epsilon}{2}} + \frac{\epsilon}{2}\frac{\partial H \left(\mathbf{w}_{t+\epsilon}, \mathbf{p}_{t+\frac{\epsilon}{2}} \right)}{\partial \mathbf{w}}.
\end{split}
\end{equation}
The discretisation errors arising from the numerical leapfrog integration necessitates the performance of a Metropolis-Hastings acceptance step, which is used to determine if the proposed sample is accepted or not \citep{neal1993bayesian, neal2011mcmc}. The HMC sampling scheme proceeds va a Gibbs sampling scheme by first sampling the momentum and then sample a new set of parameters given the drawn momentum. Algorithm \ref{alg:hmc} shows the pseudo-code for the HMC where $\epsilon$ is a discretisation step size. The leapfrog steps are repeated until the maximum trajectory length $L$ is reached.

\begin{algorithm}[!ht]
  \caption{Hamiltonian Monte Carlo}
   \label{alg:hmc}
  \begin{algorithmic}[1]
    \item[] \textbf{Input}: $N$, $\epsilon$, $L$, $w_{\text{init}}$, $H(w, p)$
    \item[] \textbf{Output}: $(w)^N_{m=0}$

    \STATE $w_0\leftarrow w_{\text{init}}$
    \FOR{$m\rightarrow 1$ \KwTo $N$}
        \STATE $p_{m-1}\sim\mathcal{N}(0,\mathbf{M})$
        \STATE $p_m$, $w_m$ = \textbf{Leapfrog}($p_{m-1}$, $w_{m-1}$, $\epsilon$, $L$, $H$)
        \STATE $\delta H = {H(w_{m-1}, p_{m-1})} - {H(w_{m}, p_{m})}$
        \STATE $\alpha_m = \min\left(1, \exp\left(\delta H\right)\right)$
        \STATE $u_m \sim $ Unif$(0,1)$
        \STATE $w_m$ = \textbf{Metropolis}($\alpha_m$, $u_m$, $w_m$, $w_{m-1}$)
    \ENDFOR
    
    \item[]
    
      \item[] \textbf{function} Leapfrog($p$, $w$, $\epsilon$, $L$, $H$)
   \FOR{$t\leftarrow 1$ \KwTo $L$}
    \STATE$p \leftarrow p+ \frac{\epsilon}{2}\frac{\partial H}{\partial w}\left(w, p \right)$
    \STATE$w\leftarrow w + \epsilon{p}$
    \STATE$ p \leftarrow p + \frac{\epsilon}{2} \frac{\partial H}{\partial w}\left(w, p\right)$
  \ENDFOR
  \item[] \textbf{return} $-p$, $w$
  
  \item[]
   \item[] \textbf{function} Metropolis($\alpha_m$, $u_m$, $w_m$, $w_{m-1}$)
    \IF{$\alpha_m < u_m$}
      \STATE $w_m = w_{m-1}$
    \ELSE
      \STATE $w_m = w_{m}$
    \ENDIF
  \item[] \textbf{return} $w_m$
  \end{algorithmic}
\end{algorithm}

An issue that we are yet to address is the selection of the covariance mass matrix $\mathbf{M}$ in HMC. The mass matrix $\mathbf{M}$ is commonly set to equal the identity matrix  $\mathbf{I}$. Although this produces good results, this approach is not always the optimal approach to use.  In the following sections we address the selection of $\mathbf{M}$ by setting $\mathbf{M}$ to be: 1) a stochastic process as well as 2) using a metric that takes into account the local geometry of the target. These two approaches give result to the Quantum-Inspired Hamiltonian Monte Carlo and Riemannian Manifold Hamiltonian Monte Carlo algorithms respectively.

\section{ Quantum-Inspired  Hamiltonian Monte Carlo}
\label{sec:qihmc}

Quantum-Inspired  Hamiltonian Monte Carlo (QIHMC) sets the covariance mass matrix $\mathbf{M}$ to be a stochastic process \citep{liu2019quantum}. Inspired by the energy-time uncertainty relation from quantum mechanics, QIHMC sets $\mathbf{M}$  to be random with a probability distribution rather than a fixed mass as is the case in HMC \citep{liu2019quantum, neal2011mcmc}. \citet{liu2019quantum} show that QIHMC leaves the target distribution invariant.

Setting the mass matrix $\mathbf{M}$ to be random alleviates HMCs inefficiency in sampling from spiky and multi-modal distributions, an example of which is the transition density of jump diffusion processes \citep{mongwe2015analysis, liu2019quantum}. In addition, the implementation of QIHMC, when compared to HMC, is straightforward: we only require an extra step of re-sampling the mass matrix \citep{liu2019quantum}. If the distribution of the mass matrix is not difficult to sample from, this will add very little overhead to the computation, but with significant gains in the target exploration \citep{liu2019quantum}. Its worth noting that HMC is a special case of QIHMC when the probability distribution over the mass matrix is the Dirac delta function \citep{liu2019quantum}.   \citet{liu2019quantum} show that QIHMC outperforms HMC, RMHMC and the No-U-Turn Sampler \citep{hoffman2014no} on a variety of spiky and multi-modal distributions which occur in sparse modeling via bridge regression, image denoising and Bayesian neural network pruning.

As the Hamiltonian in QIHMC is still separable, the dynamics are integrated using the leapfrog scheme outline in equation \eqref{eq:leapfrog} - with the only difference being that the mass matrix is chosen from a user specified  distribution for each time step. The pseudo-code for the QIHMC algorithm is shown in Algorithm \ref{alg:qihmc}. 

The issue we are yet to address is what distribution over $\mathbf{M}$ to use. This is still an open research problem \citep{liu2019quantum}. In this paper, we consider the simple case where $\mathbf{M}$ is a diagonal matrix with the entries being sampled from a log-normal distribution where the mean is zero and the variance is 1. Setting the variance to 1 worked well on all the datasets, but could be tuned to obtain better results. Alternately, one could have sampled the full covariance matrix from the Whishart distribution \citep{nabirye2018wishart}.

\begin{algorithm}[!ht]
  \caption{Quantum-Inspired Hamiltonian Monte Carlo}
   \label{alg:qihmc}
  \begin{algorithmic}[1]
    \item[] \textbf{Input}: $N$, $\epsilon$, $L$, $w_{\text{init}}$, $H(w, p)$
    \item[] \textbf{Output}: $(w)^N_{m=0}$

    \STATE $w_0\leftarrow w_{\text{init}}$
    \FOR{$m\rightarrow 1$ \KwTo $N$}
        \STATE $\mathbf{M}\sim\mathcal{P_\mathbf{M}}(\mathbf{M}) \quad \textbf{\text{ <-- only difference with HMC}}$
        \STATE $p_{m-1}\sim\mathcal{N}(0,\mathbf{M})$
        \STATE $p_m$, $w_m$ = \textbf{Leapfrog}($p_{m-1}$, $w_{m-1}$, $\epsilon$, $L$, $H$)
        \STATE $\delta H = {H(w_{m-1}, p_{m-1})} - {H(w_{m}, p_{m})}$
        \STATE $\alpha_m = \min\left(1, \exp\left(\delta H\right)\right)$
        \STATE $u_m \sim $ Unif$(0,1)$
        \STATE $w_m$ = \textbf{Metropolis}($\alpha_m$, $u_m$, $w_m$, $w_{m-1}$)
    \ENDFOR

  \end{algorithmic}
\end{algorithm}

\section{ Riemannian Manifold Hamiltonian Monte Carlo}
\label{sec:rmhmc}
As seen from the previous section, although QIHMC reduces some of the sampling inefficiencies of HMC, it introduces more hyperparameters that still need to be tuned. These hyperparameters are the distribution over the mass matrix, as well as the parameters of the distribution - which is not ideal.

Riemannian Manifold Hamiltonian Monte Carlo (RMHMC) relies on the geometry of the unormalised posterior to set the mass matrix \citep{girolami2011riemann}. The method provides a fully automated adaptation mechanism and reduces the pilot runs needed to tune hyperparameters, as is the case with QIHMC \citep{girolami2011riemann, liu2019quantum}. RMHMC uses the Riemannian structure of the parameter space and thus automatically adapts to the local manifold structure at each step based on the metric tensor \citep{girolami2011riemann}. 

One however still needs to select the metric tensor to use. In this work, we set the $\mathbf{M}$  to be equal to the Hessian of the target posterior evaluated at the current state \citep{girolami2011riemann}. For the instances where the negative log-density is highly non-convex such as jump diffusion processes, the SoftAbs metric is used to give a positive definite approximation of the expected Hessian while retaining its eigenvectors \citep{betancourt2013generalizing}.

The  Hamiltonian that lies on a Riemannian manifold is expressed by replacing $\mathbf{M}$ with $\mathbf{G}(\mathbf{w})$:
\begin{equation}
\begin{split}
    \frac{{\rm d} \mathbf{w}}{\partial t} &=  \frac{\partial {\rm H}(\mathbf{w},\mathbf{p})}{\partial \mathbf{p}} = \mathbf{G}(\mathbf{w})^{-1}\mathbf{p}  
    \\
\quad \frac{{\rm d} \mathbf{p}}{\partial t} &=  -\frac{\partial {\rm H}(\mathbf{w},\mathbf{p})}{\partial \mathbf{w}}= \nabla_w \mathbf{U}(\mathbf{w}) - \nabla_w\mathcal{N}(0,\mathbf{G}(\mathbf{w})^{-1})
\end{split}
\end{equation}
where $\mathbf{G}(\mathbf{w})$ is the Hessian of the negative log-target. Since the Hamiltonian is non-separable, the leapfrog integrator defined in equation \eqref{eq:leapfrog} is no longer applicable due to the volume conservation requirements \citep{cobb2019introducing, betancourt2013generalizing, girolami2011riemann}. The most common solution for solving the non-separable Hamiltonian equations  is to rely on the implicit generalised leapfrog algorithm \citep{cobb2019introducing, betancourt2013generalizing, girolami2011riemann}. The term implicit refers to the computationally expensive first-order implicit Euler integrators, which are  fixed-point iterations that are run until convergence \citep{cobb2019introducing, girolami2011riemann}.  \citet{cobb2019introducing} introduce an explicit symplectic integration scheme for RMHMC, which less computationally than the commonly used implicit scheme. We do not consider the explicit scheme of  \citet{cobb2019introducing} in this work.

The pseudo-code for the implicit generalised leapfrog algorithm use in this paper is shown in Algorithm \ref{alg:rmhmc}. The algorithm for RMHMC is computationally expensive \citep{cobb2019introducing, betancourt2013generalizing}. This can be seen in Algorithm \ref{alg:rmhmc} where a single step in the implicit generalised leapfrog contains two ‘while’ loops, which both require expensive update steps for each iteration \citep{cobb2019introducing, beckers1981}. This will result in long execution times for RMHMC compared to HMC and QIHMC, which both use explicit integration schemes in the form of the leapfrog integrator.

\begin{algorithm}[!ht]
  \caption{Riemannian Manifold Hamiltonian Monte Carlo}
   \label{alg:rmhmc}
  \begin{algorithmic}[1]
    \item[] \textbf{Input}: $N$, $\epsilon$, $L$, $w_{\text{init}}$, $H(w, p)$
    \item[] \textbf{Output}: $(w)^N_{m=0}$

    \STATE $w_0\leftarrow w_{\text{init}}$
    \FOR{$m\rightarrow 1$ \KwTo $N$}
        \STATE $p_{m-1}\sim\mathcal{N}(0,\mathbf{M(w_{m-1})})$
        \STATE $p_m$, $w_m$ = \textbf{Integrator}($p_{m-1}$, $w_{m-1}$, $\epsilon$, $L$, $H$)
        \STATE $\delta H = {H(w_{m-1}, p_{m-1})} - {H(w_{m}, p_{m})}$
        \STATE $\alpha_m = \min\left(1, \exp\left(\delta H\right)\right)$
        \STATE $u_m \sim $ Unif$(0,1)$
        \STATE $w_m$ = \textbf{Metropolis}($\alpha_m$, $u_m$, $w_m$, $w_{m-1}$)
    \ENDFOR
    
    \item[]

     \item[] \textbf{function} Integrator($p$, $w$, $\epsilon$, $L$, $H$)
   \FOR{$t\leftarrow 1$ \KwTo $L$}
    \STATE$p \leftarrow $ \textbf{FixedPointMomentum}($p$, $w$, $\epsilon$, $L$, $H$)
    \STATE$w\leftarrow $ \textbf{FixedPointParameters}($p$, $w$, $\epsilon$, $L$, $H$)
    \STATE $p = p +\frac{\epsilon}{2}\frac{\partial H}{\partial w}\left(w, p \right)$
  \ENDFOR
  \item[] \textbf{return} $-p$, $w$
  \item[]
     \item[] \textbf{function} FixedPointMomentum($p_0$, $w_0$, $\epsilon$, $L$, $H$)
   \STATE $p= p_0$
   \WHILE{ $\Delta p > 10^{-6}$}
   \STATE $p^{*} = p +\frac{\epsilon}{2}\frac{\partial H}{\partial w}\left(w_0, p \right)$
   \STATE $\Delta p = \text{max}_i\{|p_i -p^{*}_i|\}$
   \STATE $p = p^{*}$
   \ENDWHILE
   \item[] \textbf{return} $p$
   \item[]
   \item[] \textbf{function} FixedPointParameters($p_0$, $w_0$, $\epsilon$, $L$, $H$)
   \STATE $w = w_0$
   \WHILE{ $\Delta w > 10^{-6}$}
   \STATE $w^{*} = w_0 +\frac{\epsilon}{2}\frac{\partial H}{\partial p}\left(w_0, p \right) +\frac{\epsilon}{2}\frac{\partial H}{\partial p}\left(w, p \right)$
   \STATE $\Delta w = \text{max}_i\{|w_i -w^{*}_i|\}$
   \STATE $w = w^{*}$
   \ENDWHILE
   \item[] \textbf{return} $w$
  \end{algorithmic}
\end{algorithm}

\section{Proposed Antithetic Sampling Algorithms}
\label{sec:anti}

Following the approach presented in \citet{mongwe2020antithetic}, suppose we have two random variables $X$ and $Y$, that are not necessarily independent, with the same marginal distribution $U$, we have that:
\begin{align*}
\text{Var} \left[\frac{f(X)+f(Y)}{2} \right ] = \frac{1}{4} \left[   \text{Var} f(X) + \text{Var} f(Y)\right] 
+ \frac{1}{2} \times \text{Cov} \left [ f(X), f(Y) \right].
\end{align*}
When $f(X)$ and $f(Y)$ are negatively correlated, the average of the two random variables will produce a lower variance than the average of two independent variables \citep{mongwe2020antithetic}. This equation forms the basis for the antithetic sampling algorithms that we present in this paper. 

The full benefit of antithetic sampling is greatest when the distribution $U$ is symmetrical about some vector \citep{fishman1983antithetic,piponi2020hamiltonian, mongwe2020antithetic}. For the majority of target distribution of interest to machine learning researchers, the symmetry holds only approximately \citep{piponi2020hamiltonian, mongwe2020antithetic}. The approximate symmetry results in approximate anti-coupling, which nonetheless still provides good variance reduction \citep{piponi2020hamiltonian, mongwe2020antithetic}. 

The pseudo-code for the new antithetic algorithms that we are proposing is presented in Algorithms \ref{alg:anti-qihmc} - \ref{alg:anti-rmhmc}, with the algorithm for A-HMC shown in Algorithm \ref{alg:anti-hmc}. The difference between the antithetic algorithms and the original algorithms is that the momentum variable and the uniform random variable in the Metropolis-Hastings acceptance step are shared between the two chains.

\begin{algorithm}[!ht]
  \caption{Antithetic HMC}
   \label{alg:anti-hmc}
  \begin{algorithmic}[1]
    \item[] \textbf{Input}: $N$, $\epsilon$, $L$, $w^x_{\text{init}}$, $w^y_{\text{init}}$, $H(w, p)$
    \item[] \textbf{Output}: $(w^x)^N_{m=0}$, $(w^y)^N_{m=0}$

    \STATE $w^x_0\leftarrow w^x_{\text{init}}$
    \STATE $w^y_0\leftarrow w^y_{\text{init}}$
    \FOR{$m\rightarrow 1$ \KwTo $N$}
        \STATE $p^x_{m-1}\sim\mathcal{N}(0,\mathbf{M})$
        \STATE $p^y_{m-1} = -p^x_{m-1}$
        \STATE $p^x_m$, $w^x_m$ = \textbf{Leapfrog}($p^x_{m-1}$, $w^x_{m-1}$, $\epsilon$, $L$, $H$)
        \STATE $p^y_m$, $w^y_m$ = \textbf{Leapfrog}($p^y_{m-1}$, $w^y_{m-1}$, $\epsilon$, $L$, $H$)
        \item[]
        
        \STATE $\delta H^x = {H(w^x_{m-1}, p^x_{m-1})} - {H(w^x_{m}, p^x_{m})}$
        \STATE $\delta H^y = {H(w^y_{m-1}, p^y_{m-1})} - {H(w^y_{m}, p^y_{m})}$
        \item[]
       \STATE $\alpha^x_m = \min\left(1, \exp\left(\delta H^x\right)\right)$
        \STATE $\alpha^y_m = \min\left(1, \exp\left(\delta H^y\right)\right)$
       \item[]
        \STATE $u_m \sim $ Unif$(0,1)$
        \STATE $w^x_m$ = \textbf{Metropolis}($\alpha^x_m$, $u_m$, $w^x_m$, $w^x_{m-1}$)
        \STATE $w^y_m$ = \textbf{Metropolis}($\alpha^y_m$, $u_m$, $w^y_m$, $w^y_{m-1}$)
   \ENDFOR
      
  \end{algorithmic}
\end{algorithm}

\begin{algorithm}[!ht]
  \caption{Antithetic QIHMC}
   \label{alg:anti-qihmc}
  \begin{algorithmic}[1]
    \item[] \textbf{Input}: $N$, $\epsilon$, $L$, $w^x_{\text{init}}$, $w^y_{\text{init}}$, $H(w, p)$
    \item[] \textbf{Output}: $(w^x)^N_{m=0}$, $(w^y)^N_{m=0}$

    \STATE $w^x_0\leftarrow w^x_{\text{init}}$
    \STATE $w^y_0\leftarrow w^y_{\text{init}}$
    \FOR{$m\rightarrow 1$ \KwTo $N$}
        \STATE $\mathbf{M}\sim\mathcal{P_\mathbf{M}}(\mathbf{M})$
        \STATE $p^x_{m-1}\sim\mathcal{N}(0,\mathbf{M})$
        \STATE $p^y_{m-1} = -p^x_{m-1}$
        \STATE $p^x_m$, $w^x_m$ = \textbf{Leapfrog}($p^x_{m-1}$, $w^x_{m-1}$, $\epsilon$, $L$, $H$)
        \STATE $p^y_m$, $w^y_m$ = \textbf{Leapfrog}($p^y_{m-1}$, $w^y_{m-1}$, $\epsilon$, $L$, $H$)
        \item[]
        
        \STATE $\delta H^x = {H(w^x_{m-1}, p^x_{m-1})} - {H(w^x_{m}, p^x_{m})}$
        \STATE $\delta H^y = {H(w^y_{m-1}, p^y_{m-1})} - {H(w^y_{m}, p^y_{m})}$
        \item[]
       \STATE $\alpha^x_m = \min\left(1, \exp\left(\delta H^x\right)\right)$
        \STATE $\alpha^y_m = \min\left(1, \exp\left(\delta H^y\right)\right)$
       \item[]
        \STATE $u_m \sim $ Unif$(0,1)$
        \STATE $w^x_m$ = \textbf{Metropolis}($\alpha^x_m$, $u_m$, $w^x_m$, $w^x_{m-1}$)
        \STATE $w^y_m$ = \textbf{Metropolis}($\alpha^y_m$, $u_m$, $w^y_m$, $w^y_{m-1}$)
   \ENDFOR
      
  \end{algorithmic}
\end{algorithm}

\begin{algorithm}[!ht]
  \caption{Antithetic RMHMC}
   \label{alg:anti-rmhmc}
  \begin{algorithmic}[1]
    \item[] \textbf{Input}: $N$, $\epsilon$, $L$, $w^x_{\text{init}}$, $w^y_{\text{init}}$, $H(w, p)$
    \item[] \textbf{Output}: $(w^x)^N_{m=0}$, $(w^y)^N_{m=0}$

    \STATE $w^x_0\leftarrow w^x_{\text{init}}$
    \STATE $w^y_0\leftarrow w^y_{\text{init}}$
    \FOR{$m\rightarrow 1$ \KwTo $N$}
        \STATE $p^x_{m-1}\sim\mathcal{N}(0,\mathbf{M(w^x_{m-1})})$
        \STATE $p^y_{m-1} = -p^x_{m-1}$
        \STATE $p^x_m$, $w^x_m$ = \textbf{Integrator}($p^x_{m-1}$, $w^x_{m-1}$, $\epsilon$, $L$, $H$)
        \STATE $p^y_m$, $w^y_m$ = \textbf{Integrator}($p^y_{m-1}$, $w^y_{m-1}$, $\epsilon$, $L$, $H$)
        \item[]
        
        \STATE $\delta H^x = {H(w^x_{m-1}, p^x_{m-1})} - {H(w^x_{m}, p^x_{m})}$
        \STATE $\delta H^y = {H(w^y_{m-1}, p^y_{m-1})} - {H(w^y_{m}, p^y_{m})}$
        \item[]
       \STATE $\alpha^x_m = \min\left(1, \exp\left(\delta H^x\right)\right)$
        \STATE $\alpha^y_m = \min\left(1, \exp\left(\delta H^y\right)\right)$
       \item[]
        \STATE $u_m \sim $ Unif$(0,1)$
        \STATE $w^x_m$ = \textbf{Metropolis}($\alpha^x_m$, $u_m$, $w^x_m$, $w^x_{m-1}$)
        \STATE $w^y_m$ = \textbf{Metropolis}($\alpha^y_m$, $u_m$, $w^y_m$, $w^y_{m-1}$)
   \ENDFOR
      
  \end{algorithmic}
\end{algorithm}

\section{Experiments}
\label{sec:exp}
We test the proposed variance reduction schemes on two posterior distributions. We first study jump diffusion processes whose transition density is an infinite mixture of normal random variables, with the mixing weights being Poisson probabilities \citep{mongwe2015analysis, merton_1976}. We then study Bayesian logistic regression, which is a commonly use tool for binary classification \citep{girolami2011riemann, mongwe2020antithetic}.

\subsection{Jump diffusion Processes}

The jump diffusion model that is analysed in this paper is the \citet{merton_1976} one-dimensional Markov process $\{S_{t},t\geq 0\}$ which is characterised by the following Stochastic Differential Equation (SDE):
 \begin{equation}
d\ln S_{t}= \left(\mu-\frac{1}{2}\sigma^2\right) dt+\sigma dB_{t}+ d\left(\sum\limits_{i=1}^{N_{t}}Y_{i}\right)
\label{eq:jumpdiffusion}
\end{equation}
where $\mu$ is the drift coefficient, $\sigma$ is the diffusion coefficient, $\{B_{t},t\geq 0\}$ is a standard Brownian motion process, $Y_{i}$ is the random size of the $i$th jump  and $\{N_{t},t\geq 0\}$ is a  Poisson process  with intensity  $\lambda$. Furthermore, the  jump sizes $Y_{i}$ are assumed to be independent and identically distributed random variables, and  independent of both $\{N_{t},t\geq 0\}$ and $\{B_{t},t\geq 0\}$. In addition, it is assumed that $Y_{i}\sim N(\mu_{jump},\sigma^{2}_{jump})$.  This model is  the jump diffusion model used by  \citet{press1967} in the context of finding a model that accurately describes the underlying return process and  \citet{merton_1976} in the context of pricing options when the returns are discontinuous.

The SDE in equation (\ref{eq:jumpdiffusion}) implies a particular transition density. In order  to use the MLE approach, the transition density of the jump diffusion process has to be determined. As shown by \citet{mongwe2015analysis}, the transition density of the returns of the jump diffusion process  in equation (\ref{eq:jumpdiffusion}) is given as:
\begin{equation}
\begin{split}
\mathbb{P}(\ln S(t+\tau)=w|\ln S(t)=x) 
= \sum\limits_{n=0}^{\infty} \frac{e^{-\lambda \tau}(\lambda \tau)^n}{n!}\frac{\phi\left(\frac{w-x-(\mu\tau+n\mu_{jump})}{\sqrt{\sigma^2\tau+n\sigma^2_{jump}}}\right)}{\sqrt{\sigma^2\tau+n\sigma^2_{jump}}}
\label{eq:jumpPDF}
\end{split}
\end{equation}
where $w$ and $x$ are real numbers, $\tau$ is the time difference between  $S(t+\tau)$ and $S(t)$ and $\phi$ is the probability density function of a standard normal random variable. The transition density in equation (\ref{eq:jumpPDF}) is an infinite mixture of normal  distributions. \citet{mongwe2015analysis} shows that the likelihood function has many points where it is not defined.  These features of the density mean that calibrating this model is notoriously difficult, with maximum likelihood based approaches often being stuck in local minima \citep{mongwe2015analysis}.

\subsection{Bayesian Logistic Regression}
Bayesian logistic regression is a standard tool for binary classification that is used extensively in various disciplines. The Bayesian logistic regression model is defined as follows: 
\begin{equation}
\begin{split}
    w_d &\sim\mathcal{N}(0,1) \\
    y_n &\sim\mathcal{B}ern (\sigma(x^T_nw))
\end{split}
\end{equation}
where $w$ are the weights associated with each input feature, $\mathcal{B}ern$ is the Bernoulli distribution and $\sigma$ is the sigmoid activation function. The likelihood function is given by:
\begin{equation}
    L(w) = \sum_{i}^{N} y_i \text{log}( w^Tx_i) +(1-y_i) \text{log}(1-w^Tx_i)  
\end{equation}
where $N$ is the number of observations.

\subsection{Datasets}
The details of the datasets used in this paper are displayed in Table \ref{tab:datasets}. The datasets consists of financial market data that we use to calibrate the jump diffusion process model and real world benchmark datasets for performing classification which we model using Bayesian logistic regression.

We calibrate jump diffusion processes to two financial market datasets. These are real world datasets across different financial markets. The datasets consists of daily prices that were obtained from Google Finance. The specifics of the datasets are as follows:
\begin{itemize}
    \item \textit{S\&P 500 dataset}: Daily data for the stock index from 1 Jan 2017 to 31 Dec 2020. The prices were converted into log returns.
    \item \textit{USDZAR dataset}: Daily data for the currency from 1 Jan 2017 to 31 Dec 2020. The prices were converted into log returns.
\end{itemize}

There are three datasets that we modeled using Bayesian logistic regression. All the datatsets have two classes. The specifics of the datasets are:

\begin{itemize}
    \item \textit{Australian credit dataset} - This dataset has 14 features and 690 data points. 
    \item \textit{South African fraud dataset} - This dataset is of audit findings of South African municipalities \citep{mongwe2020efficacy, mongwe2020_survey}. The dataset has 14 features, which are financial ratios, and 1 560 data points. 
    \item \textit{German credit dataset} - This dataset has 25 features and 1 000 data points.
\end{itemize}

The features for the Bayesian logistic datasets were normalised to have mean zero and standard deviation one. The jump diffusion datasets used the time series of log returns as the input. For each dataset, the chain was run 10 times from different starting positions and the results recorded. The prior distribution over the parameters for all the datasets was Gaussian with standard deviation equal to 1. For the jump diffusion datasets, 500 samples were generated after 100 burn-in samples. For the Bayesian logistic regression, datasets 2000 samples were generated after 500 samples of burn-in.  

\begin{table} [htb]
\centering
\caption{Datasets used in this paper. BJDP is Bayesian jump diffusion process. $N$ represents the number of observations. BLR is Bayesian Logistic Regression. $D$ represents the number of model parameters.}
\label{tab:datasets}
\scalebox{1.0}{
\begin{tabular}{|c|c|c|c|c|} \hline
Dataset & Features & $N$ & Model & $D$\\ \hline
S\& 500 Index & 1 & 1 007 & BJDP & 5\\
USDZAR  & 1 & 1 425 & BJDP  & 5\\
Australian credit & 14 & 690 & BLR & 15\\
South African fraud & 14 & 1 560 & BLR & 15\\
German credit & 24 & 1 000 & BLR & 25 \\
\hline
\end{tabular}
}
\end{table}

\subsection{Performance Metrics}
The performance metrics used in this paper are the multivariate Effective Sample Size ($m\mathbb{ESS}$) and the $m\mathbb{ESS}$ normalised by the execution time. The execution time is the time taken to generate the samples after the burn-in period.  For the HMC and A-HMC algorithms we set the mass matrix $\mathbf{M} = \mathbf{I}$, which is the common approach in practice. For the RMHMC and A-RMHMC methods the hessian was used as the metric. Note that for QIHMC and its antithetic version the mass matrix $\mathbf{M}$ is stochastic.

In this work, we use the multivariate ESS methodology presented in \citet{multiESS_2019}. In this metric, the correlations between the different parameter dimensions are taken into account - which is unlike the minimum univariate ESS measure which dominates the MCMC literature\citep{girolami2011riemann, multiESS_2019, mongwe2020antithetic}. The minimum univariate ESS calculation results in the estimate of the  ESS being dominated by the parameter dimensions that mix the slowest, and ignoring all other dimensions \citep{multiESS_2019, mongwe2020antithetic}. The multivariate ESS is calculated as \citep{multiESS_2019, mongwe2020antithetic}:
$$m\mathbb{ESS} = N\times \left(\frac{|\Lambda|}{|\Sigma|} \right)^{\frac{1}{D}}$$ 
where $N$ is the number of generated samples, $D$ is the number of parameters, $\Lambda$ is the sample covariance matrix and $\Sigma$ is the estimate of the Markov chain standard error. When $D = 1$, $m\mathbb{ESS}$ is equivalent to the univariate ESS \citep{mongwe2020antithetic}.

The $m\mathbb{ESS}$  for the variance reduced chain is related to the $m\mathbb{ESS}$  of the original chain as \citep{piponi2020hamiltonian, mongwe2020antithetic}:
\begin{equation} 
\label{ess_anti}
m\mathbb{ESS}_{antithetic} = \frac{2 \times m\mathbb{ESS}_{original}}{1 + \rho}
\end{equation}
where $\rho$ is the correlation coefficient between the corresponding pairs of chains. In this paper, the correlation is taken as the maximum correlation across all the parameter dimensions, which creates a lower bound for $m\mathbb{ESS}_{antithetic}$. Equation (\ref{ess_anti}) implies that we can increase the ESS by ensuring that the correlation is as close as possible to -1, which is what the anti-coupling methodology aims to do. This also means that the $m\mathbb{ESS}_{antithetic}$ can be greater than the total number of generated samples $N$.

The number of steps used to generate each sample for HMC and QIHMC was fixed at 200, while for RMHMC it was set to 6. The step-sizes were chosen by targeting an acceptance rate of 80\% in the original chain through the dual averaging  methodology  outlined in  \citet{hoffman2014no} and \citet{mongwe2020antithetic}.

The dual averaging algorithm for HMC is shown in Algorithm \ref{alg:dual}, with the implementation for the other algorithms following similarly.  In this paper we use the same parameter settings as in Hoffman and Gelman \citep{hoffman2014no}, and these settings are shown in Algorithm \ref{alg:dual}. It is important to note that for the antithetic chains, we only adapt the step size of the original chain, and feed this step size to the other chain \citep{mongwe2020antithetic}.

\begin{algorithm}[!ht]
  \caption{HMC with dual averaging}
   \label{alg:dual}
  \begin{algorithmic}[1]
    \item[] \textbf{Input}: $N$, $w_{\text{init}}$, $\epsilon_0$, $L$, $M^{adapt}$, $\mu = \log (10 \epsilon_0)$, \\
    $\bar{\epsilon}_0 = 1$, $\bar{H}_0 =0$, $\gamma = 0.05$, $t_0 = 10$, $\kappa = 0.75$
    \item[] \textbf{Output}: $(w)^N_{m=0}$,  final step size $=\epsilon_{M^{adapt}}$

    \STATE $w_0\leftarrow w_{\text{init}}$
    \FOR{$m\rightarrow 1$ \KwTo $N$}
        \STATE $p_{m-1}\sim\mathcal{N}(0,\mathbf{M})$
        \STATE $p_m$, $w_m$ = \textbf{Leapfrog}($p_{m-1}$, $w_{m-1}$, $\epsilon_{m-1}$, $L$, $H$)
        \STATE $\delta H = {H(w_{m-1}, p_{m-1})} - {H(w_{m}, p_{m})}$
        \STATE $\alpha_m = \min\left(1, \exp\left(\delta H\right)\right)$
        
        \IF{ $m < M^{adapt}$ }
            \STATE $\bar{H}_m = \left(1- \frac{1}{ m + t_0} \right)\bar{H}_{m-1} + \frac{1}{ m + t_0}\left(\delta - \alpha_m \right)$ 
            \STATE $\log\epsilon_m =  \mu - \frac{\sqrt{m}}{\gamma} \bar{H}_m$ 
            \STATE $ \log \bar{\epsilon}_m = m^{-\kappa} \log\epsilon_m  + ( 1- m^{-\kappa}) \log \bar{\epsilon}_{m -1} $ 
        \ELSE
            \STATE $\epsilon_m = \bar{\epsilon}_{M^{adapt}}$ 
        \ENDIF
        
        \STATE $u_m \sim $ Unif$(0,1)$
        \STATE $w_m$ = \textbf{Metropolis}($\alpha_m$, $u_m$, $w_m$, $w_{m-1}$)
    \ENDFOR

  \end{algorithmic}
\end{algorithm}

\section{Results and Discussion}
\label{sec:results}
The experiments were implemented in PyTorch \citep{paszke2019pytorch} and  were carried out on a 64-bit precision CPU. In evaluating the Riemannian Manifold Hamiltonian Monte Carlo algorithms and its antithetic variant, we set a convergence tolerance of $10^{-6}$ or the completion of ten fixed point iterations. 

\begin{figure*}
    \centering
    \includegraphics[width=1.0 \textwidth]{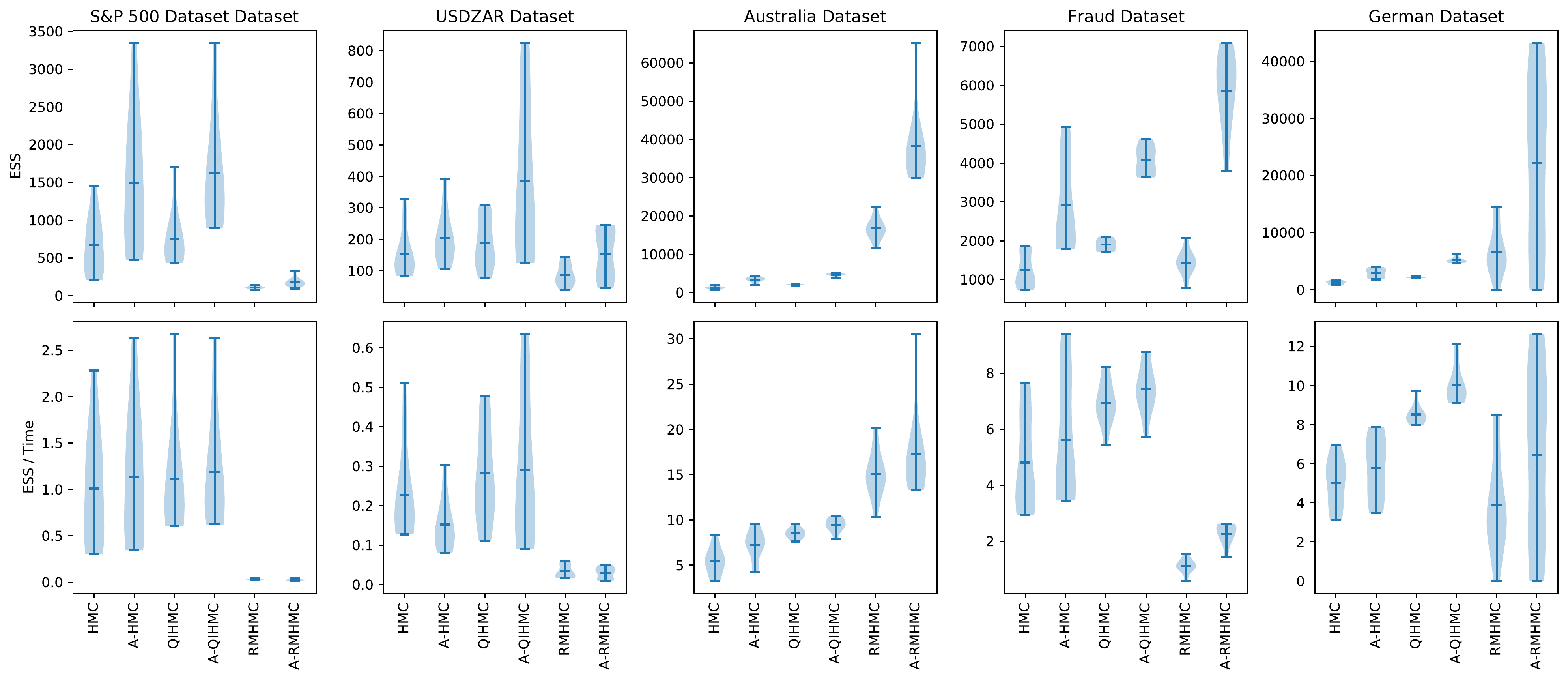}
    \caption{Results for the datasets over 10 runs of each method. For each dataset, the plots on the first row  show the multivariate effective sample size and the plots on the second row show the multivariate effective sample size normalised by execution time (in seconds). For all the plots, the larger the value the better the method. The dark horizontal line in each violin plot represents the mean value over 10 runs of each algorithm.}
    \label{fig:results}
\end{figure*}

The performance of the algorithms across different metrics is shown in Figure \ref{fig:results} and Table \ref{tab:results}. In Figure \ref{fig:results}, the plots on the first row for each dataset show the $m\mathbb{ESS}$, and the plots on the second row show the $m\mathbb{ESS}$ normalised by execution time (which is in seconds). The results are for the 10 runs of each algorithm.

As with Figure \ref{fig:results}, the execution time $t$ in Table \ref{tab:results} is in seconds. The results in Table \ref{tab:results} are the mean results over the 10 runs for each algorithm.  Note that we use the mean values over the 10 runs in Table \ref{tab:results} to form our conclusions about the performance of the algorithms. As expected, HMC has the lowest execution time $t$ on the majority of the datasets, with QIHMC being a close second. RMHMC exhibits the largest execution time across all the datasets due to the implicit nature of the numerical integration scheme.

The results show that QIHMC produces higher ESSs than HMC across all the datasets. QIHMC outperforms on an ESS basis on the jump diffusion process datasets, while RMHMC outperforms on the logistic regression datasets. This is in line with what was observed by \citet{liu2019quantum}, where QIHMC outperformed RMHMC on multi-modal and spiky distributions - jump diffusion process are an example of multi-modal and spiky distributions. Note that this is observed on both the antithetic and original versions of the methods.

A-QIHMC outperforms all the methods on a normalised ESS basis on all the datasets, except on the Australian dataset where A-RMHMC outperforms. Across all the datasets, A-HMC and its original version underperform  QIHMC and RMHMC on an ESS basis. We find that the antithetic versions of all the algorithms have higher effective sample sizes than their non-antithetic variants, which shows the usefulness of antithetic Markov Chain Monte Carlo methods.

\begin{table*}
  \centering

  \caption{Mean results over 10 runs of each algorithm. Each column represents the mean value for the specific method. For example, column one shows the mean results for HMC. The execution time $t$ is in seconds. The values in \textbf{bold} indicate that the particular method outperforms the other methods on that specific metric. For example, RMHMC outperforms in terms of $m\mathbb{ESS}$ on the Fraud dataset.}
\scalebox{1.1}{

\begin{tabular}{llllllll}

\bottomrule
 \hline
 \multicolumn{7}{|c|}{{S{\&}P 500 dataset}} \\
 \hline
 \toprule
&   HMC  & A-HMC  &    QIHMC   & A-QIHMC &    RMHMC   & A-RMHMC \\
\midrule
\midrule

$m\mathbb{ESS}$ & 667& 1 499& 758 &\textbf{1 619} & 108& 177\\
$t$ (in secs)  & \textbf{690}& 1 381& 705& 1 411& 3 481& 6 962\\
$m\mathbb{ESS}/ t$  & 1.01& 1.14&1.11 & \textbf{1.20}& 0.03& 0.03\\
\bottomrule
 \hline
 \multicolumn{7}{|c|}{{USDZAR dataset}} \\
 \hline
\toprule

$m\mathbb{ESS}$ &152 & 204&187 & \textbf{385} & 87& 154\\
$t$ (in secs)  & 680& 1 361& \textbf{666}& 1 332& 2 740& 5 480\\
$m\mathbb{ESS}/ t$  & 0.23& 0.15& 0.28& \textbf{0.30}& 0.03& 0.03\\
\bottomrule

 \hline
 \multicolumn{7}{|c|}{{Australian dataset}} \\
 \hline
\toprule

$m\mathbb{ESS}$ & 1 273& 3 406& 2 113& 4 704& 16 791 & \textbf{38 354}\\
$t$ (in secs)  & \textbf{235}& 470& 248& 496& 1 116& 2 233\\
$m\mathbb{ESS}/ t$  & 5.41 & 7.24& 8.51& 9.47& 15.04 & \textbf{17.23} \\

\bottomrule
 \hline
 \multicolumn{7}{|c|}{ {Fraud dataset}} \\
 \hline
\toprule

$m\mathbb{ESS}$ & 1 249& 2 919 & 1 901& 4 072& 1 437 & \textbf{5 866}\\
$t$ (in secs)  & \textbf{260}&  520& {275}& 551& 1 298& 2 596 \\
$m\mathbb{ESS}/ t$  & 4.81& 5.62& 6.94& \textbf{7.43} & 1.11 & 2.27 \\
\bottomrule

 \hline
 \multicolumn{7}{|c|}{{German dataset}} \\
 \hline
\toprule

$m\mathbb{ESS}$ &  1 260 & 2 910& 2 193& 5 159& 6 690& \textbf{22 182}\\
$t$ (in secs)  & \textbf{251}& 503& 257& 514& 1 708& 3 417 \\
$m\mathbb{ESS}/ t$  &  5.01& 5.79& 8.52& \textbf{10.03}& 3.91 & 6.45\\

\bottomrule
\end{tabular}
}
\label{tab:results}
\end{table*}

\section{Conclusion}
\label{sec:conclusion}

We present the antithetic versions of Riemannian Manifold Hamiltonian Monte Carlo and Quantum-Inspired Hamiltonian Monte Carlo and compare them to the antithetic version of Hamiltonian Monte Carlo on sampling from jump diffusion and Bayesian logistic regression models.

The results show that the new antithetic Riemannian Manifold Hamiltonian Monte Carlo and Quantum-Inspired Hamiltonian Monte Carlo algorithms produce higher effective sample size rates than antithetic Hamiltonian Monte Carlo. The antithetic Riemannian Manifold Hamiltonian Monte Carlo method produces the largest effective sample sizes of all the methods on the logistic regression datasets, while the antithetic Quantum-Inspired Hamiltonian Monte Carlo produces the highest effective sample size rates normalised by execution time of all the methods across all datasets and outperforms on an effective sample size basis on the jump diffusion process datasets. We find that the antithetic versions of all the algorithms have higher effective sample sizes than their non-antithetic variants, which shows the usefulness of antithetic Markov Chain Monte Carlo methods.

This work can be improved by assessing the sensitivity of the methods to the path length as, in this study, the path length was fixed and the step size tuned. In addition, tuning the distribution of the mass matrix in Quantum-Inspired Hamiltonian Monte Carlo could also improve on the results presented in this paper. A comparison of the performance of the proposed algorithms to their control-variate counterparts is also of interest.
\

\section*{Acknowledgements} \label{ac}
The work of Wilson Tsakane Mongwe and Rendani Mbuvha was supported by the Google Ph.D. Fellowships in Machine Learning. The work of Tshilidzi Marwala was supported by the National Research Foundation of South Africa. The computations in this work were performed on resources provided by the Center for High Performance Computing (CHPC) at the Council of Scientific and Industrial Research (CSIR) South Africa.

\bibliographystyle{unsrtnat}
\bibliography{references}  






\end{document}